\pdfoutput=1

\documentclass[11pt]{article}

\usepackage[preprint]{acl}

\usepackage{times}
\usepackage{latexsym}
\usepackage{multirow}

\usepackage[T1]{fontenc}

\usepackage[utf8]{inputenc}

\usepackage{microtype}

\usepackage{inconsolata}
\usepackage{tabularx}
\usepackage{graphicx}
\usepackage{booktabs}
\usepackage{subcaption}

\title{Pixels to Principles: Probing Intuitive Physics Understanding in Multimodal Language Models}

\author{
  \textbf{Mohamad Ballout}$^*$,
  \textbf{Serwan Jassim}$^*$,
  \textbf{Elia Bruni}
\\
\\
  Institute of Cognitive Science, University of Osnabrück, Osnabrück, Germany
\\
  \small{
    \textbf{Correspondence:} \href{mailto:email@domain}{mohamad.ballout@uos.de}
  }
}

\begin{document}
\maketitle
\def\thefootnote{*}\footnotetext{These authors contributed equally to this work}
\begin{abstract}

This paper presents a systematic evaluation of state-of-the-art multimodal large language models (MLLMs) on intuitive physics tasks using the GRASP and IntPhys 2 datasets. We assess the open-source models InternVL 2.5, Qwen 2.5 VL, LLaVA-OneVision, and the proprietary Gemini 2.0 Flash Thinking, finding that even the latest models struggle to reliably distinguish physically plausible from implausible scenarios. To go beyond performance metrics, we conduct a probing analysis of model embeddings, extracting intermediate representations at key processing stages to examine how well task-relevant information is preserved. Our results show that, depending on task difficulty, a critical vision-language misalignment can emerge: vision encoders successfully capture physical plausibility cues, but this information is not effectively utilized by the language model, leading to failures in reasoning. This misalignment suggests that the primary limitation of MLLMs in intuitive physics tasks is not the vision component but the ineffective integration of visual and linguistic information. Our findings highlight vision-language alignment as a key area for improvement, offering insights for future MLLMs development.


   
\end{abstract}

\section{Introduction}

With the rise of transformers~\cite{vaswani2017attention}, large-scale pre-training on extensive language datasets has transformed natural language processing (NLP). Models such as GPT~\cite{radford2018improving}, BERT~\cite{devlin2019bert}, and T5~\cite{raffel2020exploring} have demonstrated strong capabilities across various NLP tasks. Recently, multimodal large language models (MLLMs), including GPT-4~\cite{achiam2023gpt} and Gemini~\cite{gemini}, have emerged, integrating textual and visual inputs. While traditional language models have been extensively studied~\cite{chang2024survey}, MLLMs remain relatively underexplored~\cite{wang2024exploring} due to their novelty, computational demands, and evaluation complexity.

In this paper, we investigate the ability of state-of-the-art multimodal large language models (MLLMs) to reason about intuitive physics concepts. We evaluate these models using the GRASP~\cite{jassim2023grasp} and IntPhys 2~\cite{bordes2025intphys} datasets, which contain simulated videos depicting physically plausible and implausible scenarios. These scenarios test adherence to fundamental principles such as object permanence and gravity. For instance, a model should recognize that a ball must fall due to gravity rather than remain suspended in mid-air, as depicted in GRASP.

When GRASP was introduced, \citet{jassim2023grasp} reported that open-source MLLMs, such as Video-LLaMA \cite{zhang2023video} and Video-ChatGPT \cite{maaz2023video}, failed to perform better than chance on their dataset. While MLLMs have since advanced in vision-language tasks, we show that even state-of-the-art models still struggle with distinguishing plausible from implausible videos—a task where humans achieved ~80\% accuracy. Models like Qwen 2.5 VL \cite{bai2025qwen2}, InternVL 2.5 \cite{chen2024expanding}, LLaVA-OneVision \cite{li2024llava}, and Gemini 2.0 Flash Thinking fail to exceed 54\% accuracy. To understand the reasons behind this persistent failure, we not only conduct a systematic performance evaluation but also go further by probing the internal representations of these models. We extract embeddings from key processing layers and train simple classifiers to determine whether models encode task-relevant physical plausibility cues. Additionally, we use t-SNE clustering to visualize the structure of these embeddings, revealing that while vision features form distinct clusters corresponding to different physical concepts, this structure becomes less distinguishable after passing through the language model, specifically in level 2 of GRASP. This analysis reveals a critical misalignment between vision and language components: while vision encoders successfully capture distinctions between plausible and implausible scenes, this information is not effectively utilized by the language model.

The main contributions of this paper can be summarized as follows:
\begin{itemize} 
\item We systematically evaluate state-of-the-art multimodal LLMs on intuitive physics tasks using GRASP and IntPhys 2 datasets, showing that even the latest models, including Qwen 2.5 VL, InternVL 2.5, LLaVA-OneVision, and Gemini 2.0 Flash Thinking, struggle to exceed chance performance.

\item We go beyond accuracy metrics by probing model embeddings and demonstrate that, depending on task difficulty, vision encoders retain task-relevant information while language models often fail to leverage it. Additionally, t-SNE clustering reveals that vision representations naturally cluster according to physical concepts, but this structure deteriorates after alignment with the language model.

\item 
We identify vision-language misalignment as the key bottleneck, showing that information degrades as it passes through the language model, indicating poor vision-language alignment.

\end{itemize}

\section{Related Work}

As our work examines the physics reasoning capabilities of MLLMs, we review relevant literature on MLLMs, including studies and datasets used to evaluate their physics reasoning.

\subsection{Multimodal LLMs}

MLLMs emerged following the integration of image patches into transformers \cite{dosovitskiy2020image}, enabling researchers to combine similar architectures in vision and language models. Typically, an MLLM pairs a pre-trained image encoder like CLIP \cite{radford2021learning} with a pre-trained language decoder from models such as the LLaMA family \cite{touvron2023llama} or the Vicuna family \cite{chiang2023vicuna}. CLIP is a pioneering model that aligns vision and text by pre-training an image encoder using hundreds of millions of online image-text pairs.

Effective alignment between vision and language remains essential and is typically achieved by introducing trainable connectors that bridge visual and language features.
Several models follow this approach, including Flamingo \cite{alayrac2022flamingo} and BLIP-2 \cite{li2023blip}. In our experiments, we evaluate InternVL 2.5~\cite{chen2024expanding}, Qwen 2.5 VL~\cite{bai2025qwen2}, and LLaVA-OneVision~\cite{li2024llava}, all of which integrate a pre-trained Vision Transformer (ViT)~\cite{dosovitskiy2020image} encoder with their respective LLMs via a connection layer. These models are trained in two main stages: first, pre-training for feature alignment, where the LLM and ViT remain frozen while the connector is trained (except in Qwen 2.5 VL, where the ViT is also trained); second, end-to-end fine-tuning of all model layers. Training data consists of matched image-text pairs and visual question-answering datasets.

Current proprietary models like GPT-4o~\footnote{\url{https://openai.com/index/hello-gpt-4o}} (OpenAI) and Gemini~\footnote{\url{https://deepmind.google/technologies/gemini}} (Google) also incorporate vision components. However, their closed-source nature restricts detailed implementation insights and prevents analysis of latent features. As of this writing, only Gemini officially supports video prompting. While GPT-4o can process video frames, our initial tests reveal highly inaccurate performance with frequent hallucinations.


Several image and video datasets have been developed to evaluate models' understanding of physics. Prior to the emergence of MLLMs, datasets such as VQA \cite{antol2015vqa}, CLEVR \cite{johnson2017clevr}, and CLEVRER \cite{yi2020clevrer} were designed to study visual commonsense reasoning. As MLLMs have evolved, more challenging datasets have been introduced. For instance, PIQA \cite{bisk2020piqa} evaluates physics commonsense reasoning in language models using images, while VEC \cite{li2023can} assesses fundamental physics understanding by focusing on concepts like mass, temperature, and hardness. iVISPAR~\cite{mayer2025ivispar} introduces an interactive benchmark that highlights MLLMs' strengths and limitations in visual-spatial reasoning. While some studies \cite{liu2022things,zhang2022visual} suggest MLLMs can acquire spatial commonsense knowledge, others \cite{liu2023visual} indicate they struggle with understanding object orientations and relationships.

In addition, several video datasets have been developed to evaluate models' intuitive physics reasoning abilities. Benchmarks such as Physical Concepts \cite{piloto2022intuitive}, IntPhys \cite{riochet2021intphys}, IntPhys 2 \cite{bordes2025intphys}, InfLevel \cite{weihs2022benchmarking}, AVoE \cite{dasgupta2021avoe, dasgupta2021benchmark}, and GRASP \cite{jassim2023grasp} were designed for this purpose. \citet{weihs2022benchmarking} evaluated ten different vision models on infant-level physics tasks and found that they performed only slightly above chance. Similarly, \citet{jassim2023grasp} demonstrated that even advanced MLLMs struggled with intuitive physics tasks, performing at chance levels (50\%), whereas human performance reached approximately 80\%. Furthermore, \citet{garrido2025intuitivephysicsunderstandingemerges} found that MLLMs are significantly outperformed by self-supervised vision models across the IntPhys, GRASP, and InfLevel benchmarks, which is in line with the findings in IntPhys 2~\cite{bordes2025intphys}.

Building on recent advancements in MLLMs and the release of new models, we further investigate the GRASP~\cite{jassim2023grasp} and IntPhys 2 ~\cite{bordes2025intphys} datasets using newer models and additional techniques. Specifically, we evaluate MLLMs on GRASP and IntPhys 2 by directly analyzing their responses and probing latent features. We selected these two datasets because prior MLLMs performed around chance level on them, and they provide the most comprehensive and diverse sets of intuitive physics videos.

\section{Datasets}\label{dataset}

GRASP~\cite{jassim2023grasp} and IntPhys 2~\cite{bordes2025intphys} are benchmarks that can be used to evaluate intuitive physics understanding in multimodal language models (MLLMs). GRASP is structured into two levels: basic visual understanding (Level 1) and intuitive physics (Level 2). For Level 1, we use four scenes covering concepts such as shape, color, directionality, and movement, with each concept comprising 128 videos. Level 2 comprises 16 scenes categorized by physics concepts such as collision, continuity, gravity, inertia, object permanence, solidity, and unchangeableness, each including 128 plausible and 128 implausible videos. IntPhys 2 extends intuitive physics evaluation through 1,416 photorealistic videos, covering four core principles: object permanence, object immutability, spatio-temporal continuity, and solidity.  In the main set, videos are arranged into easy, medium, and hard difficulty levels, with each scenario consisting of paired plausible and implausible events designed to probe models using a violation-of-expectation framework. Given that previous MLLM performed around chance level, these datasets facilitate an in-depth analysis of limitations in physics reasoning capabilities.

\section{MLLM Evaluation}
\citet{jassim2023grasp} showed that open-source MLLMs, such as Video-LLaMA~\cite{zhang2023video}, PandaGPT~\cite{su2023pandagpt}, VTimeLLM~\cite{huang2024vtimellm}, and Video-ChatGPT~\cite{maaz2023video}, performed no better than chance on their GRASP dataset. They hypothesized that this might be due to the relatively small size (7B and 13B parameters) of these models compared to state-of-the-art LLMs, which are typically an order of magnitude larger. The recent release of larger MLLMs enables further investigation of this hypothesis.
At the same time, these newer models have demonstrated significantly improved results on vision-language benchmarks, such as MathVista~\cite{lu2024mathvista}, MMMU~\cite{yue2023mmmu}, and MMBench~\cite{liu2024mmbench}, which motivates a reassessment of MLLM performance on GRASP, where we expect to see improvements as well. To strengthen our conclusions, we additionally evaluate these models on the newly introduced IntPhys 2 benchmark, which complements GRASP and enables more generalizable insights.

\begin{figure*}[th]
    \centering
    \begin{subfigure}[b]{0.48\textwidth}
        \centering
        \includegraphics[width=\textwidth]{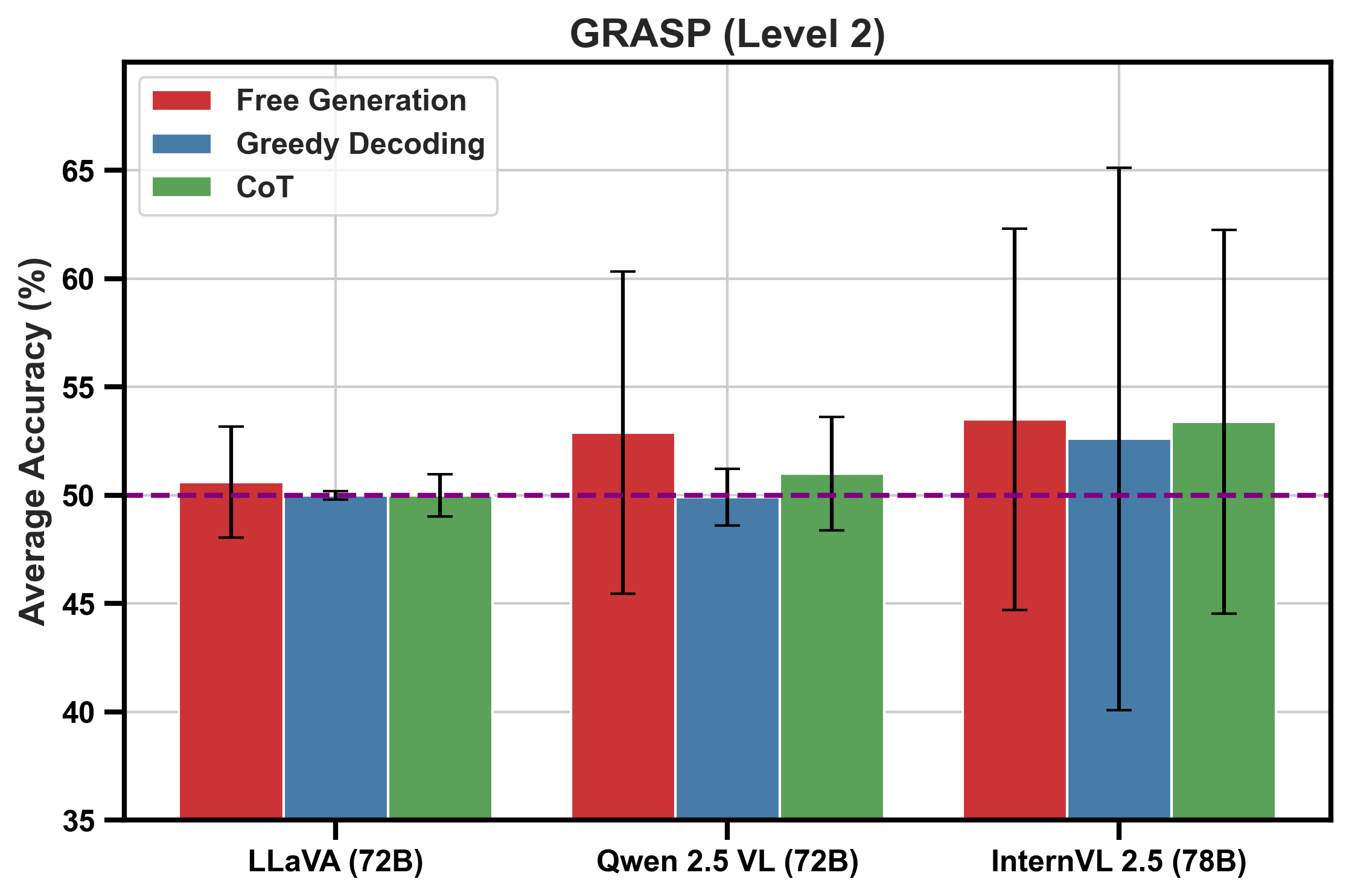}
    \end{subfigure}
    \hfill
    \begin{subfigure}[b]{0.48\textwidth}
        \centering
        \includegraphics[width=\textwidth]{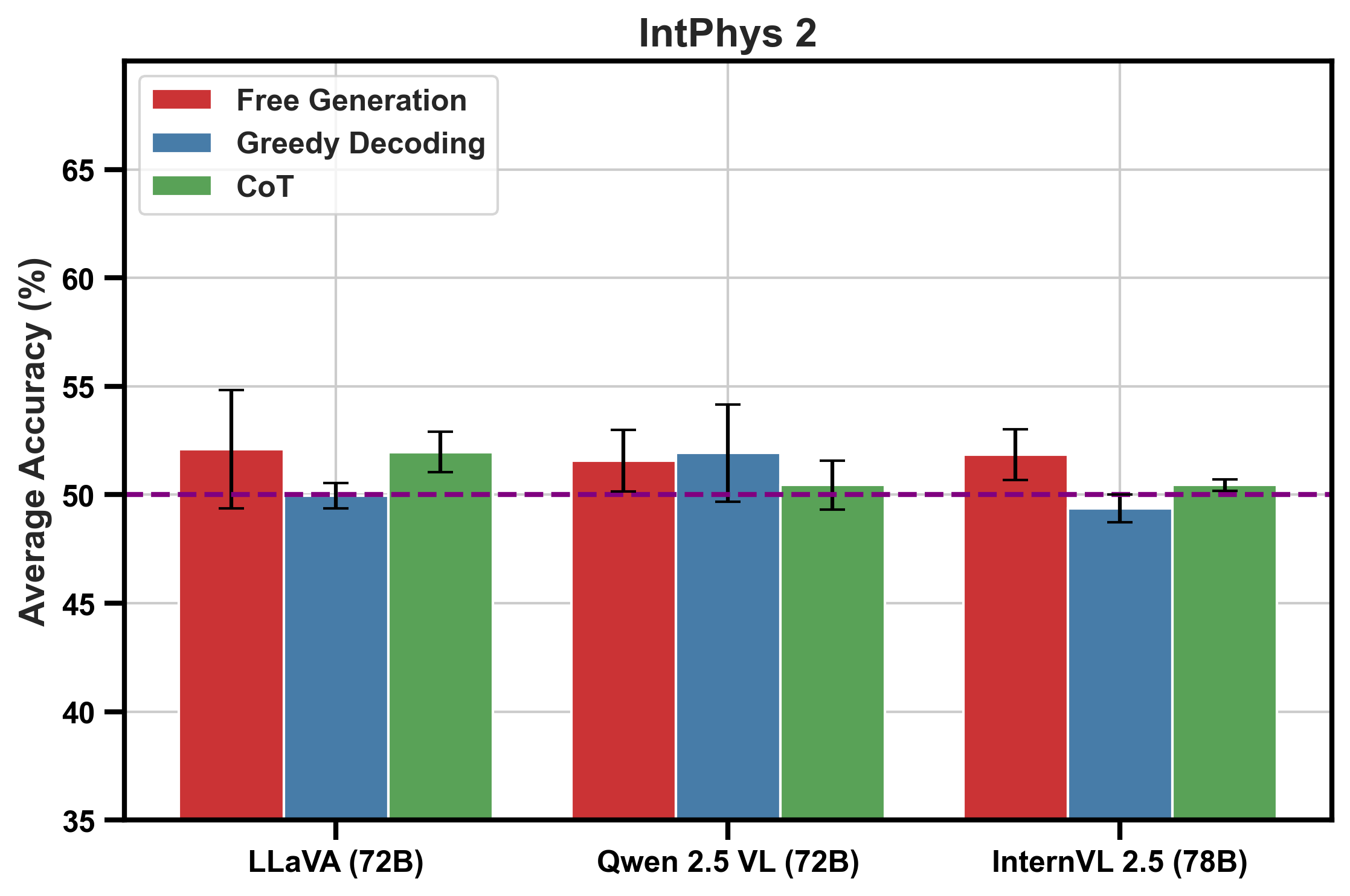}
    \end{subfigure}
    \caption{Average accuracy (\%) across the intuitive physics tests of GRASP and IntPhys 2 for the largest models. The dashed line represents chance performance. Error bars denote the standard deviation across all tests of the respective dataset.}
    \label{fig:vlm-results-large}
\end{figure*}

\subsection{Intuitive Physics Understanding}
\label{intres}

In this work, we assess the current performance of MLLMs on GRASP and IntPhys 2 using three different model families: LLaVA-OneVision (0.5B, 7B, 72B) \cite{li2024llava}, Qwen 2.5 VL (3B, 7B, 72B) \cite{bai2025qwen2}, and InternVL 2.5 (2B, 8B, 78B) \cite{chen2024expanding}. The models are evaluated by using a prompt-video pair made up of a physically plausible or implausible video of a scene, along with the corresponding question from the dataset. For example, for the Gravity scene in GRASP, the question is:
\begin{quote}
        The video you're seeing was generated by a simulator. Given how objects behave on earth, is the trajectory of the ball plausible? Your answer should be based on the events in the video and ignore the quality of the simulation. Answer only with yes or no.
\end{quote}
IntPhys 2 uses a similar prompt; however, it does not contain any scene-specific information like the prompts from GRASP.

We evaluate the models using three different methods:
\begin{enumerate}
    \item \textbf{Free generation}: The models generate responses freely, and we parse their output to identify ``yes'' and ``no'' tokens.
    \item \textbf{Greedy decoding}: Given the probabilistic nature of LLM text generation, we directly compare the logits of the ``yes'' and ``no'' tokens, selecting the token with the higher logit.
    \item \textbf{Chain-of-Thought (CoT)}: We first prompt the model with ``What do you see in this video?'' and allow it to generate freely. In a second step, we ask the actual question about the video's plausibility and apply greedy decoding.
\end{enumerate}

For the free generation and CoT methods, we average results over three runs with different random seeds. We do not apply this procedure to greedy decoding, as it does not involve sampling from the output distribution. For all models, we use a temperature of 0.6, top-p of 0.9, and top-k of 40. Videos are sampled at 1 FPS, meaning that since each video is 10 seconds long, 10 frames are extracted.

\begin{table*}[t]
    \begin{center}
    \scalebox{0.7}{
    \begin{tabular}{l|ccc|ccc|ccc}
        \toprule
        \multirow{2}{*}{\textbf{Scene
        }} & \multicolumn{3}{c|}{\textbf{LLaVA (72B)}} & \multicolumn{3}{c|}{\textbf{Qwen 2.5 VL (72B)}} & \multicolumn{3}{c}{\textbf{InternVL 2.5 (78B)}} \\
        & free & greedy & CoT & free & greedy & CoT & free & greedy & CoT \\
        \midrule
        \midrule
        Color & \textbf{100.0} & \textbf{100.0} & \textbf{100.0} & \textbf{100.0} & \textbf{100.0} & \textbf{100.0} &  \textbf{100.0} & \textbf{100.0} & \textbf{100.0}  \\
        Directionality & 27.9 & 27.0 & 44.9 & 52.5 & \textbf{99.4} & 41.7  &  25.8 & 46.3 & 24.4    \\
        Movement & 62.0 & 91.4  & 85.4 & 97.7 & \textbf{99.2} & 99.0 &  66.7 & 67.2 & 49.7   \\
        Shape & 59.1 & 63.3 & 83.3 & 99.7 & \textbf{100.0} & \textbf{100.0}  &  83.1 & 89.1 &  \textbf{100.0} \\
        \midrule
        Average & 62.8 & 68.1 & 76.4 & 83.7 & \textbf{99.7} & 80.4 &  66.9 &  74.8 &  66.4 \\
        \bottomrule
    \end{tabular}
    }
    \end{center}
    \caption{Accuracy (\%) for the largest models on GRASP's Level 1 using free-form generation, greedy decoding and CoT prompting. For the Movement and Shape tests, chance performance corresponds to 50\%, while for Directionality and Color  it is 25\%. We highlight the best results per test.}
    \label{tab:level1_results}
\end{table*}

Results for the largest open-source models are shown in Figure~\ref{fig:vlm-results-large} and for the smaller models of the tested families in Figure~\ref{fig:vlm-results-small-models}. Most models perform at or below chance level on GRASP, but Qwen 2.5 VL and InternVL 2.5 exceed 53\% accuracy in their largest configurations. Although this outperforms smaller models, the improvement remains insufficient to support the hypothesis by \citet{jassim2023grasp} that increasing model size alone enables solving the task.
This is consistent with the results observed on IntPhys 2, where the largest models slightly outperform the smaller models, reaching up to 52\% accuracy, except for InternVL 2.5, where the 78B model performs even slightly worse than the 8B variant.
Models comparable in size to those evaluated in GRASP show only modest improvements over chance (e.g., Qwen 2.5 VL (3B) reaching 51.5\% with CoT). Overall, the gains on GRASP are notably smaller than those observed in other vision-language benchmarks—for instance, Qwen 2.5 VL (78B) improved over PandaGPT (13B) by nearly 50 percentage points on the OpenCompass leaderboard~\footnote{\url{https://rank.opencompass.org.cn/leaderboard-multimodal}}.

In addition to the slight overall improvement over models evaluated in GRASP, we observe a greater variance in performance across individual GRASP scenes, as shown by the error bars in Figure~\ref{fig:vlm-results-large}. At the scene level, Qwen 2.5 VL (7B) achieves 68.9\% on the Gravity-Inertia scene using free generation, while InternVL 2.5 (78B) reaches 97.7\% on Gravity-Support with CoT prompting.
The small standard deviation for all models in IntPhys 2, where tests are grouped based on difficulty, indicates that this benchmark is generally harder than GRASP.

We also present descriptions generated by the largest models for a GRASP video in Table~\ref{tab:descriptions}. In this simple video—where an orange ball rolls up a plank without slowing down—all models either exhibit hallucinations or fail to capture key aspects of the scene. These frequent hallucinations and omissions, observed consistently across multiple videos and models, suggest a misalignment between the visual and language modalities in current MLLMs. In light of these qualitative observations, the models' poor quantitative performance is less surprising.

Beyond open-source models, we also evaluate the proprietary state-of-the-art model Gemini 2.0 Flash Thinking on GRASP, which achieves 53.7\% accuracy—indicating substantial room for improvement.

\subsection{Elementary Visual Understanding}
Due to the poor performance of MLLMs on intuitive physics tasks, the GRASP benchmark introduced a Level 1 that focuses on basic visual understanding. 
These videos follow the same format as the intuitive physics videos but test for lower-level properties such as object colors, shapes, and simple dynamics. 
Since the models evaluated in this work continue to perform poorly on intuitive physics, we also assess them on Level 1.
Poor performance on these tasks would further support the hypothesis of misalignment between vision and language modalities.

As in the previous section, we evaluate models using free-form generation, greedy decoding, and CoT prompting. The corresponding prompts are provided in Appendix \ref{prompt1}. Table~\ref{tab:level1_results} shows the results for all Level 1 tests across the largest models.

All models achieve significantly above-chance performance on all tests with at least one evaluation method. Remarkably, Qwen 2.5 VL (72B) achieves an average accuracy of 99.7\% using greedy decoding. This strong performance on basic visual understanding tasks suggests that the models are well-aligned with respect to low-level visual properties.
In contrast, their poor performance on intuitive physics tasks in both GRASP and IntPhys 2 indicates that higher-level alignment and reasoning remain limited. Qualitative analysis suggests that these failures are due to the models overlooking critical details necessary for understanding physical plausibility.

\section{Embeddings Probing}
Given the overall poor performance of all evaluated MLLMs, we investigate the underlying reasons. Specifically, we analyze their internal representations by extracting embeddings from different model layers and training linear probes to determine whether they encode information that distinguishes physically plausible from implausible videos.

\subsection{Embeddings Extraction}

We selected several critical positions for embedding extraction, focusing primarily on the vision-language projection layer, a crucial module responsible for aligning visual features with language model embeddings. For simplicity, we refer to this layer as the \textit{projection layer}, despite variations in naming across different models. Some models, such as InternVL 2.5 and LLaVA-OneVision, specifically dedicate a separate training phase to optimize this layer while freezing other parameters. Given the importance of this alignment step, we extracted embeddings from two positions: (1) before the projection layer, specifically from the last vision transformer block, and (2) after the projection layer, at the end of the entire vision encoder.

To obtain a unified representation from the extracted embeddings, we averaged over the sequence dimension, condensing multiple token-level embeddings into a single vector per video. This transformation reduces a high-dimensional feature tensor into a compact embedding, creating a global representation of the input by averaging across spatial-temporal dimensions. The resulting vector retains the vision encoder’s feature dimensionality, ensuring a consistent and expressive output for classification. The embedding dimensions for each model are listed in Table \ref{tab:embedding-configs} Appendix \ref{embed} .

Additionally, we extracted embeddings from the final layer of the language model (before the output MLP) under two different prompting conditions to assess vision-language alignment more comprehensively. The first prompt was a simple query asking whether the video was physically plausible. The second was a detailed prompt structured as follows: 
\begin{quote}
        Determine whether the video is physically plausible or implausible based on these concepts. Collision: plausible if a rolling ball hits another and both move; implausible otherwise. Gravity: plausible if objects \dots
\end{quote}
Each concept was illustrated with examples demonstrating adherence to or violations of physical laws. The complete prompt is included in Appendix \ref{prompt}. 

\subsection{Classifier}

GRASP Level 1 includes four tasks relevant to our analysis: detecting colors, movements, directions, and shapes, with 128 videos per task. Movements and shapes are binary classification tasks, while colors and directions involve four classes.

GRASP Level 2 has 4,096 videos, evenly split across 16 distinct scenes, each with 128 plausible and 128 implausible examples. Similarly, IntPhys 2 includes 1,012 videos (506 plausible and 506 implausible). For probing, we reformulated these datasets as binary classification tasks (plausible vs. implausible).

For all datasets, we trained one-hidden-layer MLP probes (hidden size=512, ReLU activation, no dropout) for 500 epochs using cross-entropy loss, with data splits of 80\% training, 10\% validation, and 10\% testing. Hyperparameters were optimized via grid search, and reported results are averages over three random seeds.

\begin{figure*}[th]
    \centering
    \begin{subfigure}[b]{0.32\textwidth}
        \centering
        \includegraphics[width=\textwidth]{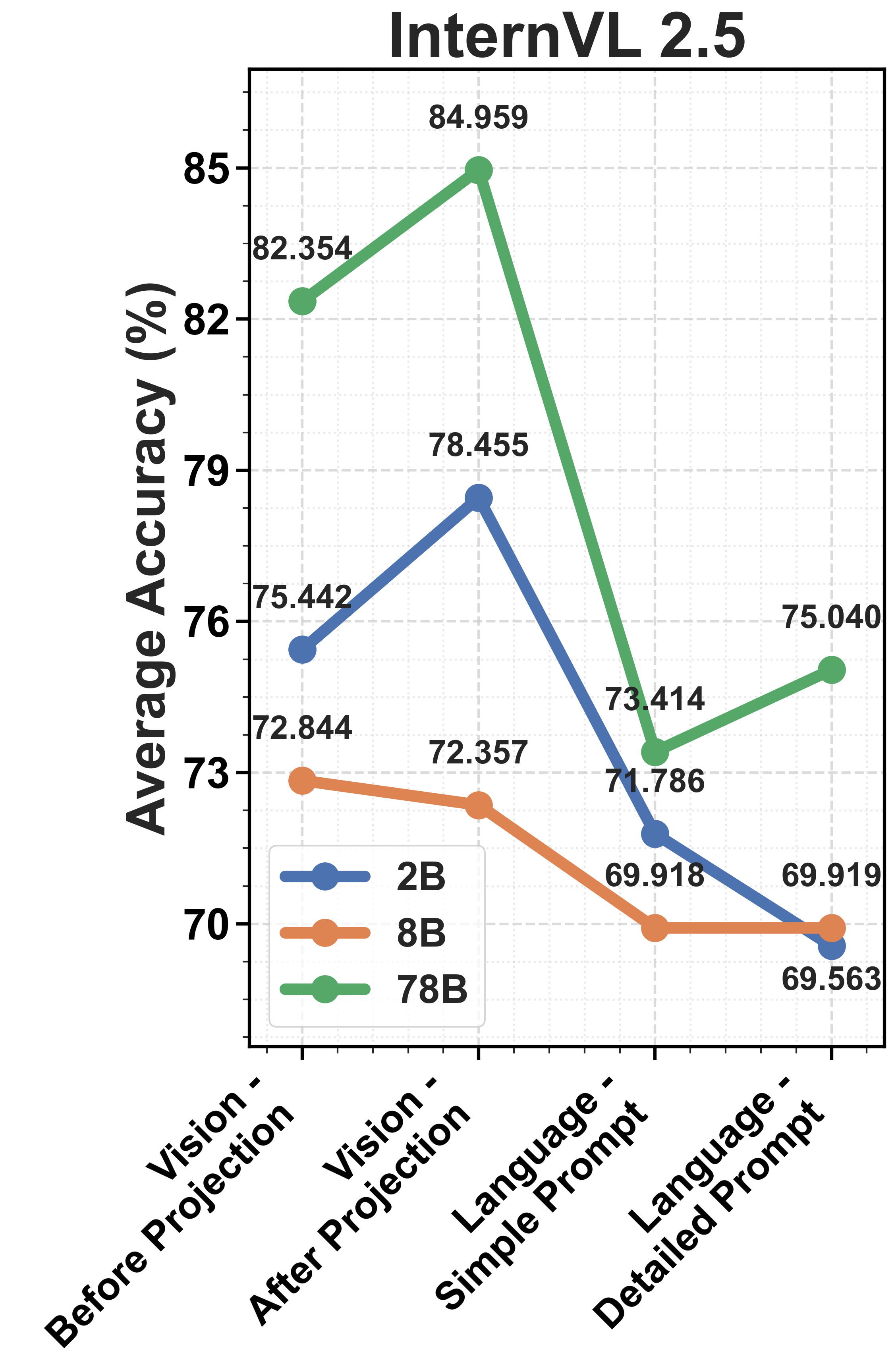}
    \end{subfigure}
    \hfill
    \begin{subfigure}[b]{0.32\textwidth}
        \centering
        \includegraphics[width=\textwidth]{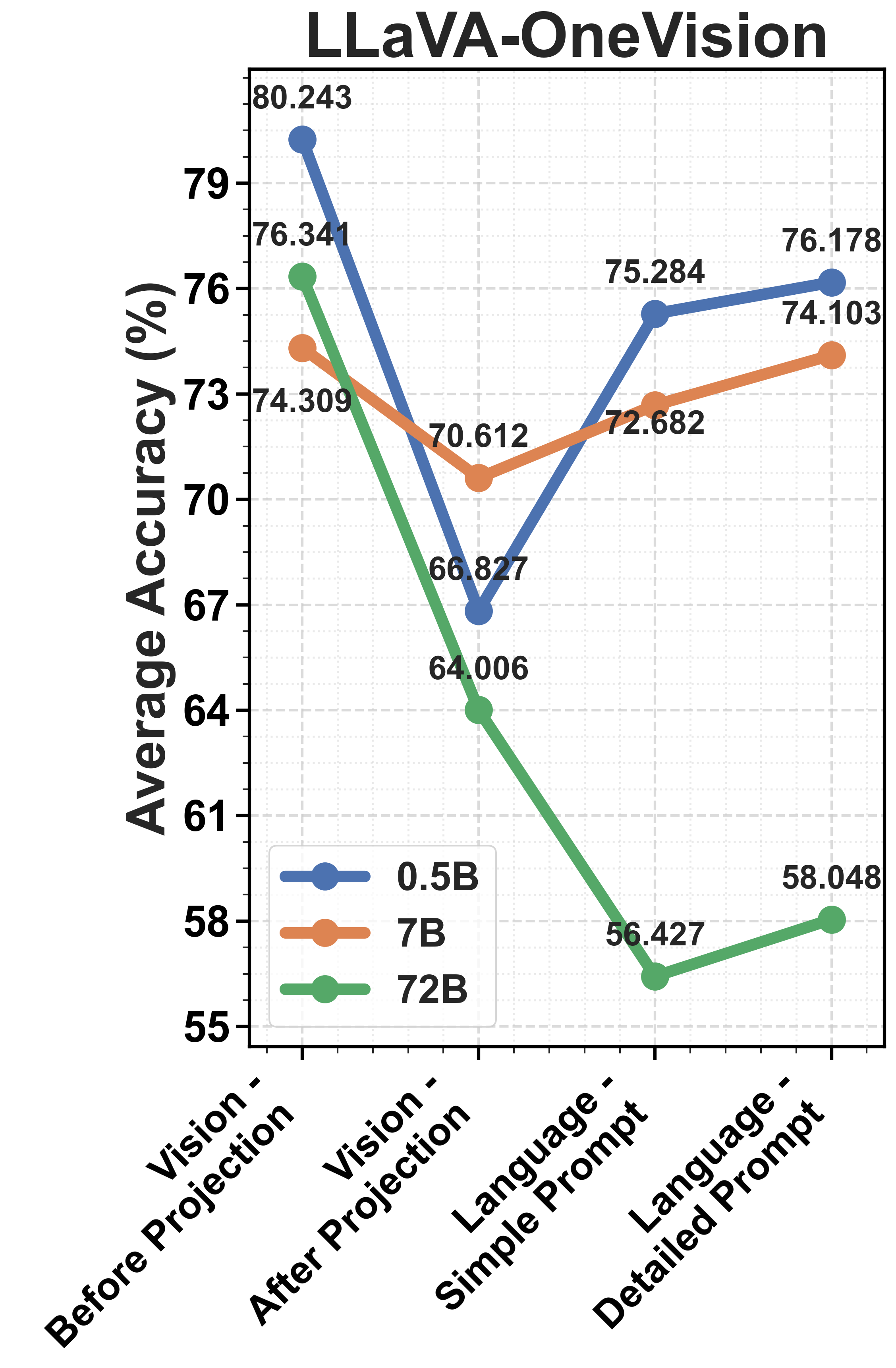}
    \end{subfigure}
    \hfill
    \begin{subfigure}[b]{0.32\textwidth}
        \centering
        \includegraphics[width=\textwidth]{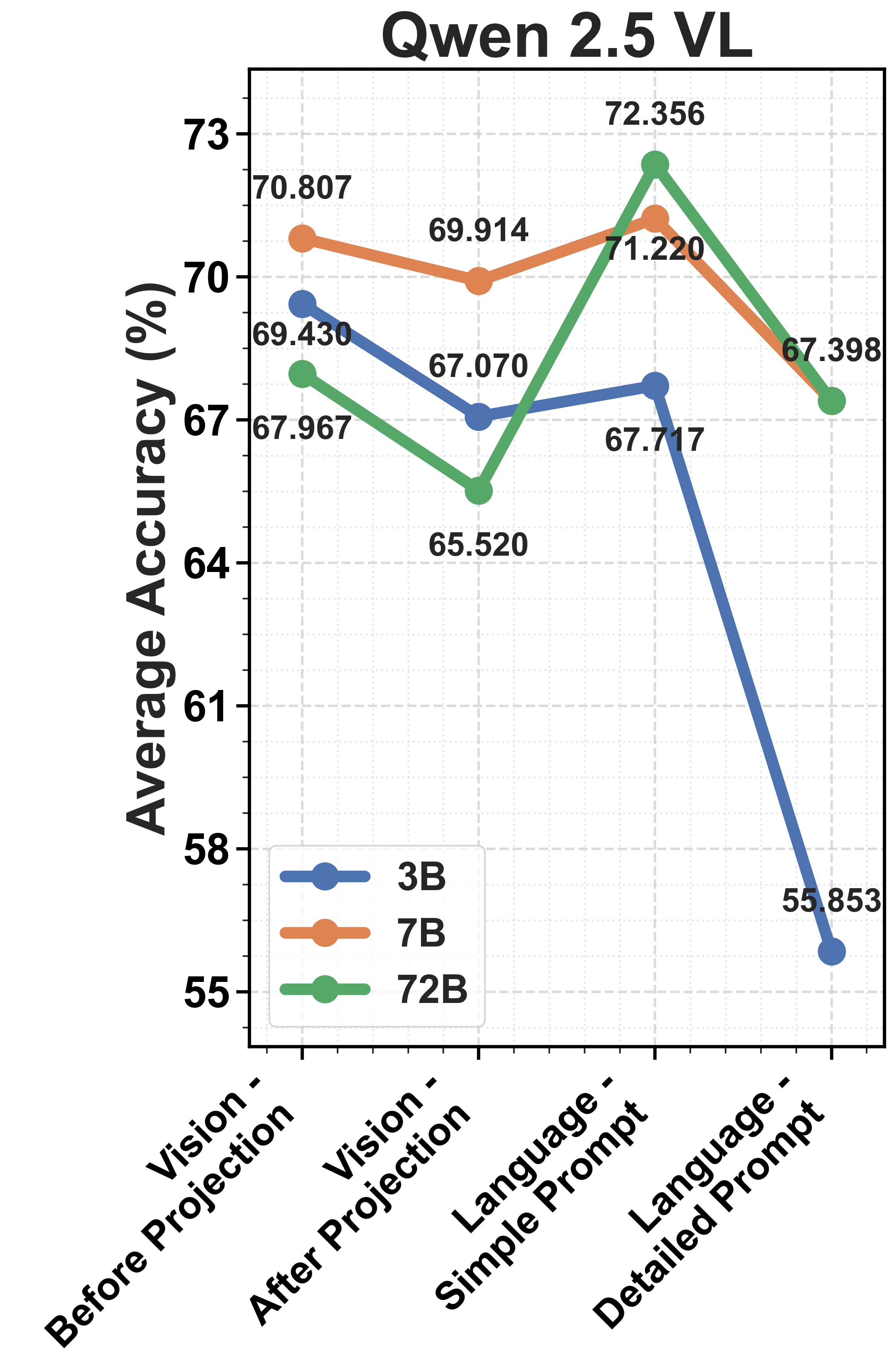}
    \end{subfigure}
    \caption{Comparison of model performance across three model families. Each subplot shows average accuracies across all GRASP scenes for the respective model sizes (InternVL 2.5: 2B, 8B, 78B; LLaVA-OneVision: 0.5B, 7B, 72B; Qwen 2.5 VL: 3B, 7B, 72B) across four embedding categories: Vision - Before Projection, Vision - After Projection, Language - Simple Prompt, and Language - Detailed Prompt.}    \label{fig:model_performance}
    \label{fig:model_performance}
\end{figure*}

\section{Results and Discussion}

In this section, we present the findings from our analysis of various embedding sets, demonstrating their ability to encode information that distinguishes between plausible and implausible videos. 

\paragraph{Insight 1: Embedding distinguishability varies with task difficulty.} For example, on Level 1 of the GRASP dataset, all probed features extracted from various positions (vision encoder or language encoder) across all tested models achieved 100\% accuracy.  For Level 1, we only tested shape and movement because directionality and color are divided into 4 classes each, leaving only 32 samples per class for training, validation, and testing. Nevertheless, the perfect accuracy on shape and movement suggests that models effectively encode basic visual properties such as shape, and movement. In contrast, on the more challenging dynamic IntPhys 2 dataset, as shown in Section~\ref{intres} and intended by the dataset's creators, all probed features from both vision and language encoders performed at chance level, demonstrating that neither component captures meaningful information regarding physical plausibility. On Level 2 of GRASP, which represents an intermediate difficulty between GRASP Level 1 and IntPhys 2, we observe variation in performance depending on the model and the embeddings location, as shown in Figure~\ref{fig:model_performance}. \textbf{Therefore, the following insights are specifically derived from analyses on Level 2 of the GRASP dataset, which contains 4,096 videos spanning 8 intuitive physics concepts across 16 distinct scenes.}

\paragraph{Insight 2: Embeddings encode sufficient information to assess video plausibility.} The results reveal that all tested embeddings contain sufficient information to determine the plausibility of videos above chance level. For example, vision embeddings from models such as InternVL 2.5 (78B) achieve an accuracy of 85.0\%, followed by LLaVA-OneVision (0.5B) at 80.2\%, and Qwen 2.5 VL (7B) at 70.8\%. These scores are significantly higher than the random baseline of 0.5, indicating that the embeddings capture meaningful distinctions. This suggests that the encoded representations are effective for assessing video plausibility.

\begin{figure*}[th]
    \centering
    \begin{subfigure}[b]{0.45\textwidth}
        \centering
        \includegraphics[width=\textwidth]{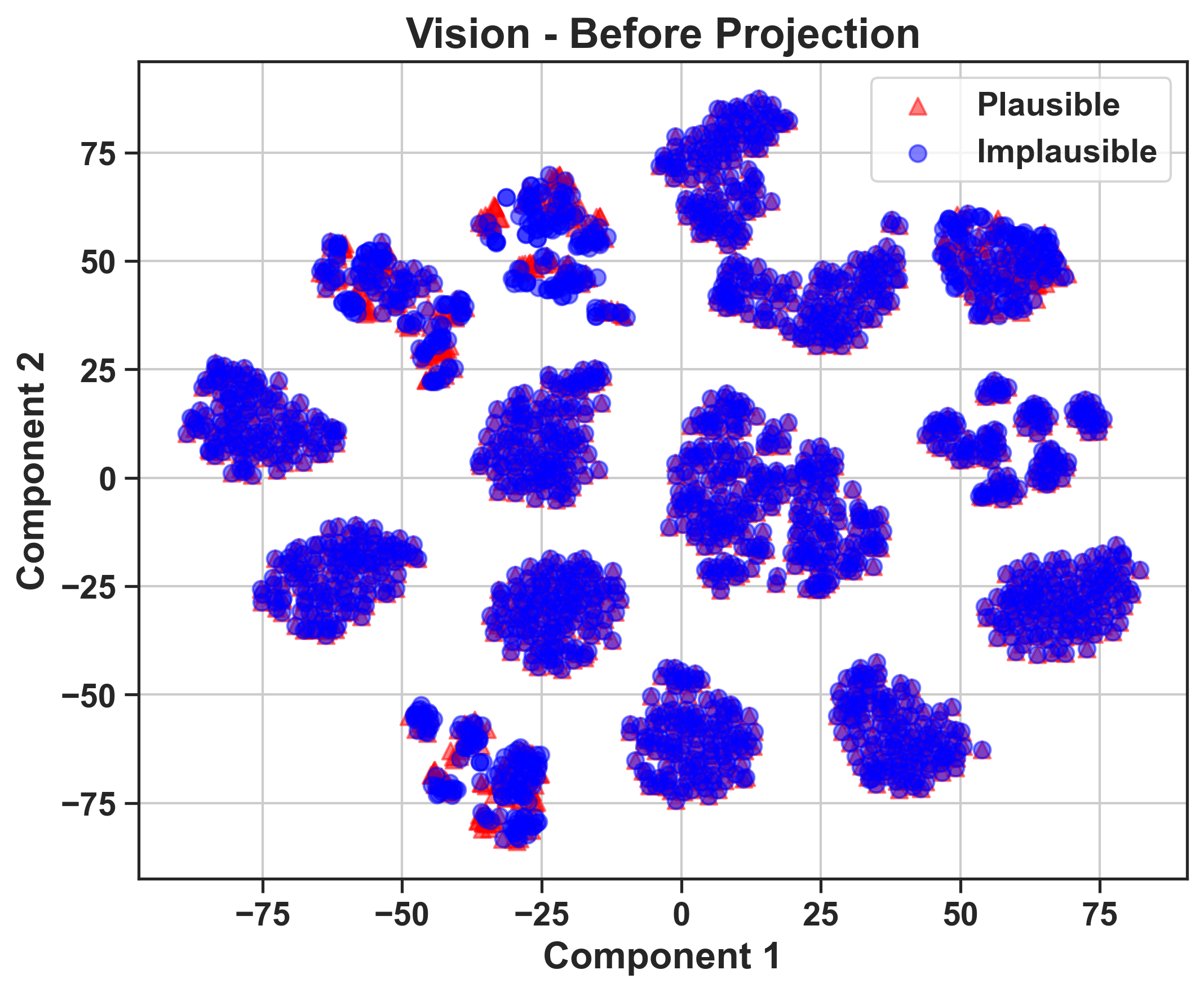} 
    \end{subfigure}
    \hfill
    \begin{subfigure}[b]{0.45\textwidth}
        \centering
        \includegraphics[width=\textwidth]{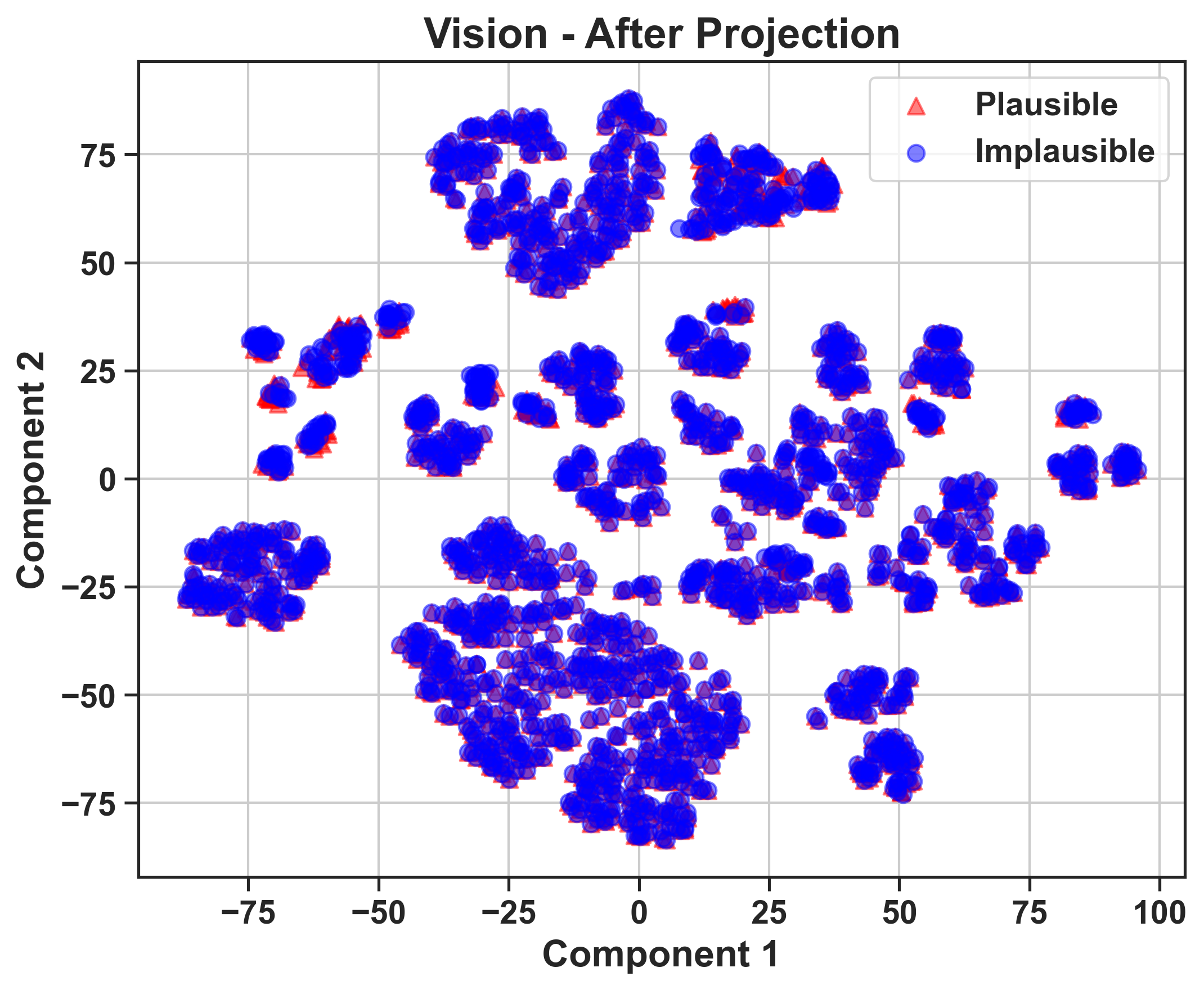} 
    \end{subfigure}
    
    \begin{subfigure}[b]{0.45\textwidth}
        \centering
        \includegraphics[width=\textwidth]{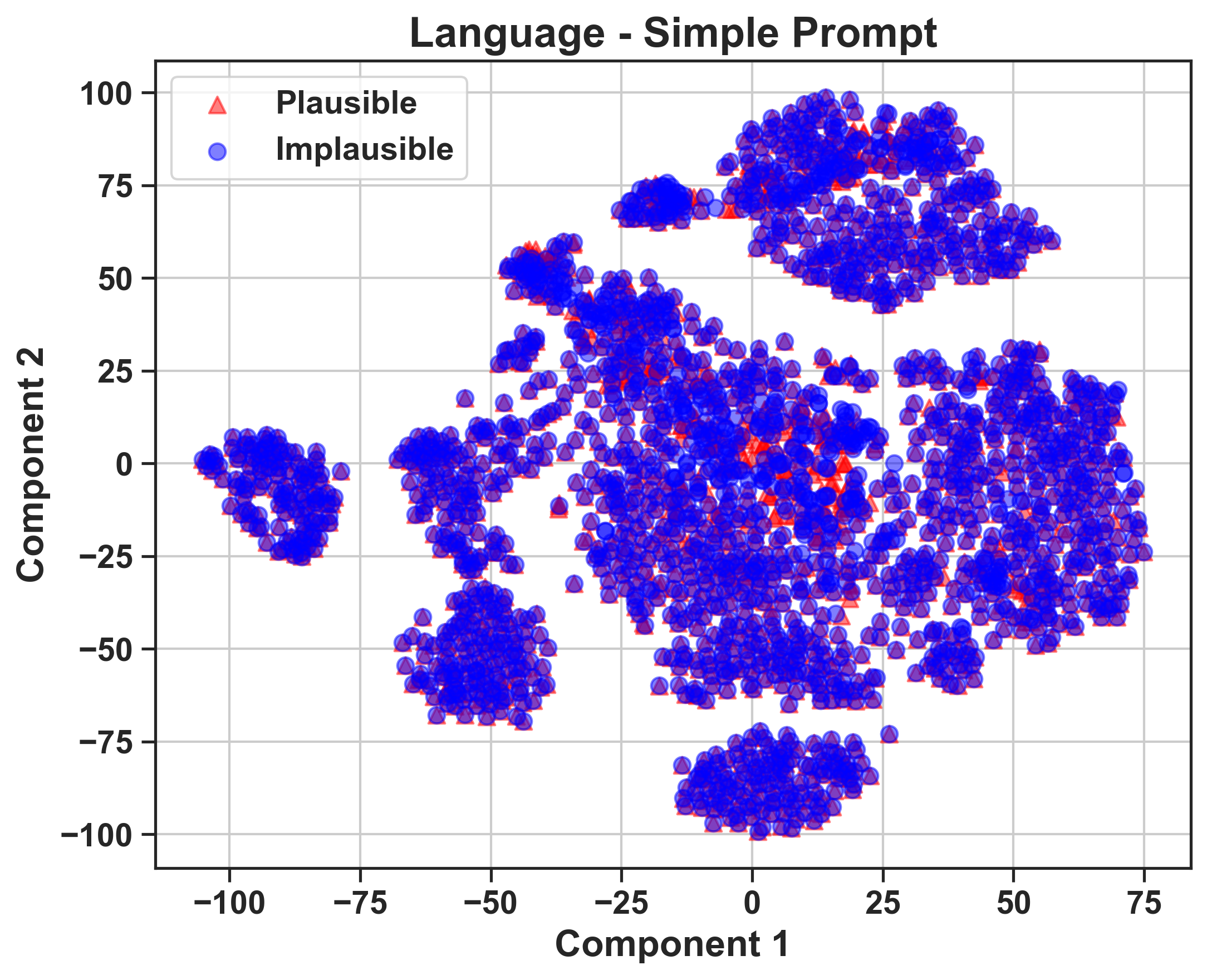} 
    \end{subfigure}
    \hfill
    \begin{subfigure}[b]{0.45\textwidth}
        \centering
        \includegraphics[width=\textwidth]{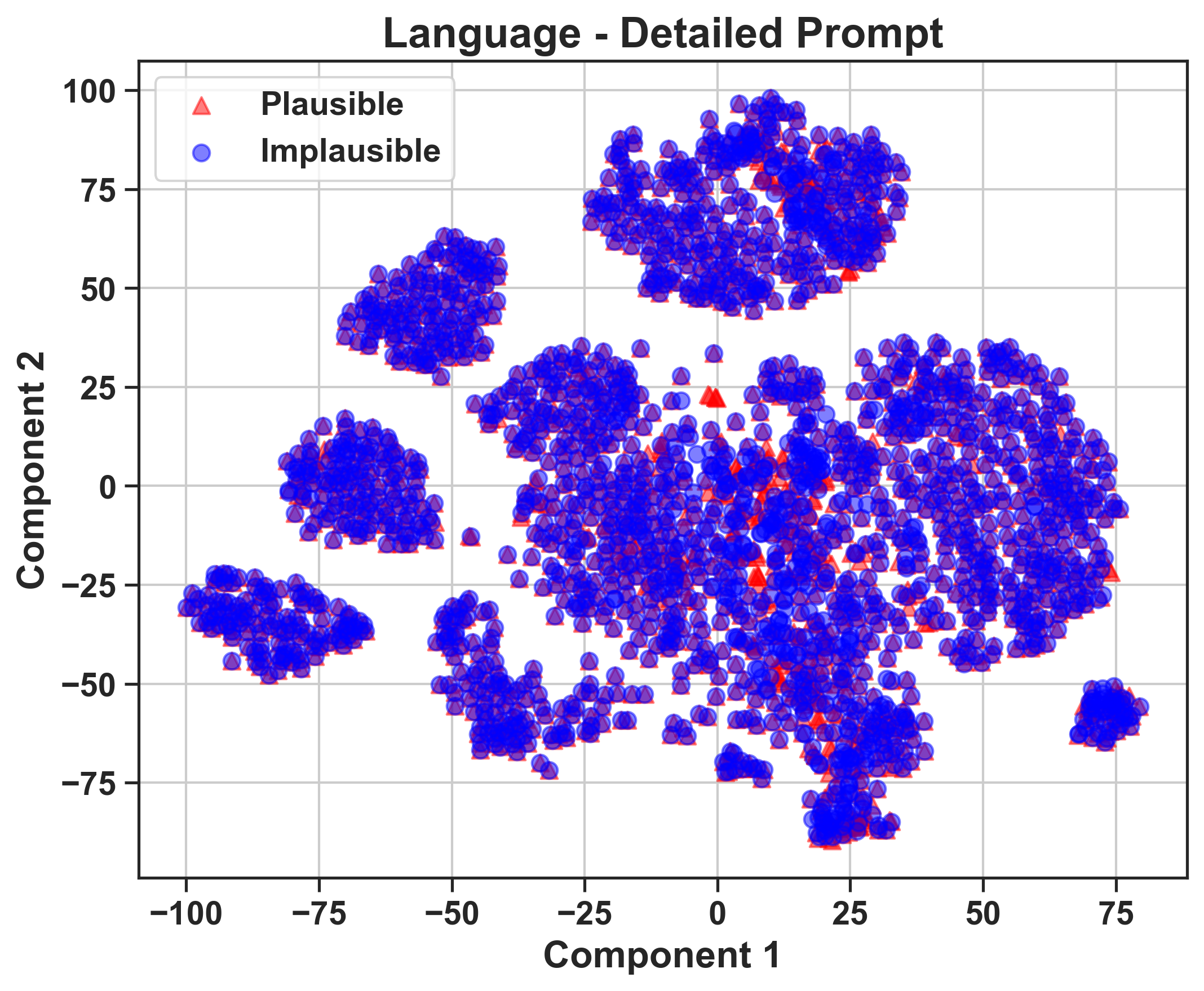} 
    \end{subfigure}
    
    \caption{2D t-SNE visualizations of video features (GRASP, level 2)  for  LLaVA-OneVision-72B across different modalities and prompting strategies: Vision - Before Projection, Vision - After Projection, Language - Simple Prompt, and Language - Detailed Prompt. Each plot distinguishes between plausible (red triangles) and implausible (blue circles) video features.}
    \label{fig:visualizations}
\end{figure*}

\paragraph{Insight 3: The language model is not able to extract relevant information from the vision embeddings.} A key observation from the results is that the vision encoder is not the limiting factor in extracting meaningful representation to distinguish between plausible and implausible videos; instead, the bottleneck lies in the language model's inability to fully extract relevant information from the vision embeddings. Comparing vision encoder performance—measured before the projection layer—against language model performance using embeddings processed with the detailed prompt reveals a consistent pattern: all nine models, irrespective of their size, perform better when probing vision embeddings. This disparity highlights a critical shortfall in the vision-to-language alignment process, suggesting that the language model fails to effectively utilize the rich information provided by the vision encoder.

With proper vision-language alignment, the task should become trivial given the explicit information provided in the detailed prompt. However, the observed performance underscores a clear alignment gap. For example, when assessing the concept of gravity, the model only needed to recognize that the ball did not fall completely to the ground to correctly classify the video as implausible—an example explicitly provided in the prompt: ``Gravity: plausible if a dropped ball falls to the floor, implausible if it stops mid-air.''

\paragraph{Insight 4: The detailed prompt has a mixed effect on model performance.} The detailed prompt, which explicitly describes relevant physical laws and specifies when a video should be classified as plausible or implausible, yielded mixed results across models. Specifically, this prompt serves as a probe for vision-language alignment by clearly defining visual conditions (e.g., if the ball remains suspended mid-air, the video is implausible). In this alignment probing task, InternVL 2.5 (78B) and all LLaVA-OneVision models performed better than with the simpler prompt, suggesting they benefit from additional explanatory context. Conversely, all Qwen 2.5 VL models performed worse, suggesting weaker alignment capabilities for this task.

\paragraph{Insight 5: The projection layer generally has a negative effect, with some exceptions.} The projection layer, which aligns visual embeddings with language model embeddings, generally reduces performance across most evaluated models. Specifically, this alignment step consistently lowers accuracy, suggesting that the projection process often degrades the quality of information derived from the vision encoder for this task. However, notable exceptions include InternVL 2.5 (2B) and InternVL 2.5 (78B), where the projection layer improves performance.

\paragraph{Insight 6: t-SNE visualizations of vision embeddings exhibit clear clustering, whereas language embeddings appear more dispersed.} To analyze clustering behavior, we applied t-SNE dimensionality reduction to visualize embeddings, as shown in Figure~\ref{fig:visualizations}. The extracted embeddings do not distinctly separate into ``plausible'' and ``implausible'' classes, likely due to the complexity of underlying patterns, which t-SNE reduces to two dimensions. However, vision embeddings before projection clearly cluster by scene, with each of the 16 dataset scenes corresponding to a specific physics concept. Figure~\ref{fig:visualizations}(a) reveals 15 distinct clusters, with one larger cluster merging two similar scenes likely due to their inherent similarity. This clustering effect gradually diminishes after embeddings pass through the projection layer, whether using simple or detailed prompts. This suggests that vision embeddings become diluted when projected and aligned to the language model space.

\section{Conclusion}

In this paper, we showed that while state-of-the-art multimodal large language models (MLLMs)—both open-source and proprietary—have slightly improved on intuitive physics tasks, they still struggle significantly. Even advanced models like Gemini Flash 2.0 Thinking, InternVL 2.5, and Qwen 2.5 VL fail to exceed 54\% accuracy when classifying the physical plausibility of a video. To understand the underlying limitations, we extracted and analyzed embeddings from open-source models at key positions within their architectures to identify potential bottlenecks.

Our analysis revealed that vision encoders consistently encode sufficient information to distinguish plausible from implausible videos, with probes trained on vision embeddings from models like InternVL 2.5 reaching 85\% accuracy. However, we identified a clear bottleneck in vision-language alignment: probes on vision embeddings outperform those on language embeddings, even when the language model is prompted in detail. This indicates that language components struggle to reason about intuitive physics, limiting overall model performance.

Future research could conduct a layer-by-layer analysis of both vision encoders and language decoders to further investigate these bottlenecks. Understanding why models that excel in general tasks perform poorly on basic intuitive physics reasoning could inform better vision-language alignment strategies, ultimately improving MLLM performance in physics-based reasoning tasks.

\section{Limitations}
This paper investigates why state-of-the-art MLLMs struggle with intuitive physics tasks. While our findings suggest that the performance bottleneck lies in vision-language alignment rather than the vision encoder, they do not fully pinpoint the exact cause of this limitation. Although vision embeddings appear to retain sufficient information, how this information is lost or misrepresented during alignment requires further investigation.

Additionally, our analysis was limited to open-source models due to the inability to access the features of Gemini and other closed-source models. As a result, assessing the specific reasons behind their failures remains challenging.
\bibliography{custom}

\begin{thebibliography}{41}
\providecommand{\natexlab}[1]{#1}

\bibitem[{Achiam et~al.(2023)Achiam, Adler, Agarwal, Ahmad, Akkaya, Aleman, Almeida, Altenschmidt, Altman, Anadkat et~al.}]{achiam2023gpt}
Josh Achiam, Steven Adler, Sandhini Agarwal, Lama Ahmad, Ilge Akkaya, Florencia~Leoni Aleman, Diogo Almeida, Janko Altenschmidt, Sam Altman, Shyamal Anadkat, et~al. 2023.
\newblock Gpt-4 technical report.
\newblock \emph{arXiv preprint arXiv:2303.08774}.

\bibitem[{Alayrac et~al.(2022)Alayrac, Donahue, Luc, Miech, Barr, Hasson, Lenc, Mensch, Millican, Reynolds et~al.}]{alayrac2022flamingo}
Jean-Baptiste Alayrac, Jeff Donahue, Pauline Luc, Antoine Miech, Iain Barr, Yana Hasson, Karel Lenc, Arthur Mensch, Katherine Millican, Malcolm Reynolds, et~al. 2022.
\newblock Flamingo: a visual language model for few-shot learning.
\newblock \emph{Advances in neural information processing systems}, 35:23716--23736.

\bibitem[{Antol et~al.(2015)Antol, Agrawal, Lu, Mitchell, Batra, Lawrence~Zitnick, and Parikh}]{antol2015vqa}
Stanislaw Antol, Aishwarya Agrawal, Jiasen Lu, Margaret Mitchell, Dhruv Batra, C~Lawrence~Zitnick, and Devi Parikh. 2015.
\newblock Vqa: Visual question answering.
\newblock In \emph{Proceedings of the IEEE International Conference on Computer Vision}, pages 2425--2433.

\bibitem[{Bai et~al.(2025)Bai, Chen, Liu, Wang, Ge, Song, Dang, Wang, Wang, Tang et~al.}]{bai2025qwen2}
Shuai Bai, Keqin Chen, Xuejing Liu, Jialin Wang, Wenbin Ge, Sibo Song, Kai Dang, Peng Wang, Shijie Wang, Jun Tang, et~al. 2025.
\newblock Qwen2. 5-vl technical report.
\newblock \emph{arXiv preprint arXiv:2502.13923}.

\bibitem[{Bisk et~al.(2020)Bisk, Zellers, Gao, Choi et~al.}]{bisk2020piqa}
Yonatan Bisk, Rowan Zellers, Jianfeng Gao, Yejin Choi, et~al. 2020.
\newblock Piqa: Reasoning about physical commonsense in natural language.
\newblock In \emph{Proceedings of the AAAI conference on artificial intelligence}, volume~34, pages 7432--7439.

\bibitem[{Bordes et~al.(2025)Bordes, Garrido, Kao, Williams, Rabbat, and Dupoux}]{bordes2025intphys}
Florian Bordes, Quentin Garrido, Justine~T Kao, Adina Williams, Michael Rabbat, and Emmanuel Dupoux. 2025.
\newblock Intphys 2: Benchmarking intuitive physics understanding in complex synthetic environments.
\newblock \emph{arXiv preprint arXiv:2506.09849}.

\bibitem[{Chang et~al.(2024)Chang, Wang, Wang, Wu, Yang, Zhu, Chen, Yi, Wang, Wang et~al.}]{chang2024survey}
Yupeng Chang, Xu~Wang, Jindong Wang, Yuan Wu, Linyi Yang, Kaijie Zhu, Hao Chen, Xiaoyuan Yi, Cunxiang Wang, Yidong Wang, et~al. 2024.
\newblock A survey on evaluation of large language models.
\newblock \emph{ACM Transactions on Intelligent Systems and Technology}, 15(3):1--45.

\bibitem[{Chen et~al.(2024)Chen, Wang, Cao, Liu, Gao, Cui, Zhu, Ye, Tian, Liu et~al.}]{chen2024expanding}
Zhe Chen, Weiyun Wang, Yue Cao, Yangzhou Liu, Zhangwei Gao, Erfei Cui, Jinguo Zhu, Shenglong Ye, Hao Tian, Zhaoyang Liu, et~al. 2024.
\newblock Expanding performance boundaries of open-source multimodal models with model, data, and test-time scaling.
\newblock \emph{arXiv preprint arXiv:2412.05271}.

\bibitem[{Chiang et~al.(2023)Chiang, Li, Lin, Sheng, Wu, Zhang, Zheng, Zhuang, Zhuang, Gonzalez et~al.}]{chiang2023vicuna}
Wei-Lin Chiang, Zhuohan Li, Zi~Lin, Ying Sheng, Zhanghao Wu, Hao Zhang, Lianmin Zheng, Siyuan Zhuang, Yonghao Zhuang, Joseph~E Gonzalez, et~al. 2023.
\newblock Vicuna: An open-source chatbot impressing gpt-4 with 90\%* chatgpt quality.
\newblock \emph{See https://vicuna. lmsys. org (accessed 14 April 2023)}, 2(3):6.

\bibitem[{Dasgupta et~al.(2021{\natexlab{a}})Dasgupta, Duan, Ang~Jr, Lin, Wang, Baillargeon, and Tan}]{dasgupta2021benchmark}
Arijit Dasgupta, Jiafei Duan, Marcelo~H Ang~Jr, Yi~Lin, Su-hua Wang, Ren{\'e}e Baillargeon, and Cheston Tan. 2021{\natexlab{a}}.
\newblock A benchmark for modeling violation-of-expectation in physical reasoning across event categories.
\newblock \emph{arXiv preprint arXiv:2111.08826}.

\bibitem[{Dasgupta et~al.(2021{\natexlab{b}})Dasgupta, Duan, Ang~Jr, and Tan}]{dasgupta2021avoe}
Arijit Dasgupta, Jiafei Duan, Marcelo~H Ang~Jr, and Cheston Tan. 2021{\natexlab{b}}.
\newblock Avoe: a synthetic 3d dataset on understanding violation of expectation for artificial cognition.
\newblock \emph{arXiv preprint arXiv:2110.05836}.

\bibitem[{Devlin et~al.(2019)Devlin, Chang, Lee, and Toutanova}]{devlin2019bert}
Jacob Devlin, Ming-Wei Chang, Kenton Lee, and Kristina Toutanova. 2019.
\newblock Bert: Pre-training of deep bidirectional transformers for language understanding.
\newblock In \emph{Proceedings of the 2019 Conference of the North American Chapter of the Association for Computational Linguistics: Human Language Technologies, Volume 1 (Long and Short Papers)}, pages 4171--4186.

\bibitem[{Dosovitskiy et~al.(2020)Dosovitskiy, Beyer, Kolesnikov, Weissenborn, Zhai, Unterthiner, Dehghani, Minderer, Heigold, Gelly et~al.}]{dosovitskiy2020image}
Alexey Dosovitskiy, Lucas Beyer, Alexander Kolesnikov, Dirk Weissenborn, Xiaohua Zhai, Thomas Unterthiner, Mostafa Dehghani, Matthias Minderer, Georg Heigold, Sylvain Gelly, et~al. 2020.
\newblock An image is worth 16x16 words: Transformers for image recognition at scale.
\newblock \emph{arXiv preprint arXiv:2010.11929}.

\bibitem[{Garrido et~al.(2025)Garrido, Ballas, Assran, Bardes, Najman, Rabbat, Dupoux, and LeCun}]{garrido2025intuitivephysicsunderstandingemerges}
Quentin Garrido, Nicolas Ballas, Mahmoud Assran, Adrien Bardes, Laurent Najman, Michael Rabbat, Emmanuel Dupoux, and Yann LeCun. 2025.
\newblock \href {https://arxiv.org/abs/2502.11831} {Intuitive physics understanding emerges from self-supervised pretraining on natural videos}.
\newblock \emph{Preprint}, arXiv:2502.11831.

\bibitem[{{Gemini Team}(2024)}]{gemini}
Google {Gemini Team}. 2024.
\newblock Gemini 1.5: Unlocking multimodal understanding across millions of tokens of context.
\newblock \emph{arXiv preprint arXiv:2403.05530}.

\bibitem[{Huang et~al.(2024)Huang, Wang, Chen, Song, and Zhu}]{huang2024vtimellm}
Bin Huang, Xin Wang, Hong Chen, Zihan Song, and Wenwu Zhu. 2024.
\newblock Vtimellm: Empower llm to grasp video moments.
\newblock In \emph{Proceedings of the IEEE/CVF Conference on Computer Vision and Pattern Recognition}, pages 14271--14280.

\bibitem[{Jassim et~al.(2023)Jassim, Holubar, Richter, Wolff, Ohmer, and Bruni}]{jassim2023grasp}
Serwan Jassim, Mario Holubar, Annika Richter, Cornelius Wolff, Xenia Ohmer, and Elia Bruni. 2023.
\newblock Grasp: A novel benchmark for evaluating language grounding and situated physics understanding in multimodal language models.
\newblock \emph{arXiv preprint arXiv:2311.09048}.

\bibitem[{Johnson et~al.(2017)Johnson, Hariharan, van~der Maaten, Fei-Fei, Lawrence~Zitnick, and Girshick}]{johnson2017clevr}
Justin Johnson, Bharath Hariharan, Laurens van~der Maaten, Li~Fei-Fei, C~Lawrence~Zitnick, and Ross Girshick. 2017.
\newblock Clevr: A diagnostic dataset for compositional language and elementary visual reasoning.
\newblock In \emph{Proceedings of the IEEE Conference on Computer Vision and Pattern Recognition}, pages 2901--2910.

\bibitem[{Li et~al.(2024)Li, Zhang, Guo, Zhang, Li, Zhang, Zhang, Zhang, Li, Liu et~al.}]{li2024llava}
Bo~Li, Yuanhan Zhang, Dong Guo, Renrui Zhang, Feng Li, Hao Zhang, Kaichen Zhang, Peiyuan Zhang, Yanwei Li, Ziwei Liu, et~al. 2024.
\newblock Llava-onevision: Easy visual task transfer.
\newblock \emph{arXiv preprint arXiv:2408.03326}.

\bibitem[{Li et~al.(2023{\natexlab{a}})Li, Li, Savarese, and Hoi}]{li2023blip}
Junnan Li, Dongxu Li, Silvio Savarese, and Steven Hoi. 2023{\natexlab{a}}.
\newblock Blip-2: Bootstrapping language-image pre-training with frozen image encoders and large language models.
\newblock In \emph{International conference on machine learning}, pages 19730--19742. PMLR.

\bibitem[{Li et~al.(2023{\natexlab{b}})Li, Xu, Dong, Zheng, Liu, Kong, and Sun}]{li2023can}
Lei Li, Jingjing Xu, Qingxiu Dong, Ce~Zheng, Qi~Liu, Lingpeng Kong, and Xu~Sun. 2023{\natexlab{b}}.
\newblock Can language models understand physical concepts?
\newblock \emph{arXiv preprint arXiv:2305.14057}.

\bibitem[{Liu et~al.(2023)Liu, Emerson, and Collier}]{liu2023visual}
Fangyu Liu, Guy Emerson, and Nigel Collier. 2023.
\newblock Visual spatial reasoning.
\newblock \emph{Transactions of the Association for Computational Linguistics}, 11:635--651.

\bibitem[{Liu et~al.(2022)Liu, Yin, Feng, and Zhao}]{liu2022things}
Xiao Liu, Da~Yin, Yansong Feng, and Dongyan Zhao. 2022.
\newblock Things not written in text: Exploring spatial commonsense from visual signals.
\newblock In \emph{Proceedings of the 60th Annual Meeting of the Association for Computational Linguistics (Volume 1: Long Papers)}, pages 2365--2376.

\bibitem[{Liu et~al.(2024)Liu, Duan, Zhang, Li, Zhang, Zhao, Yuan, Wang, He, Liu et~al.}]{liu2024mmbench}
Yuan Liu, Haodong Duan, Yuanhan Zhang, Bo~Li, Songyang Zhang, Wangbo Zhao, Yike Yuan, Jiaqi Wang, Conghui He, Ziwei Liu, et~al. 2024.
\newblock Mmbench: Is your multi-modal model an all-around player?
\newblock In \emph{European conference on computer vision}, pages 216--233. Springer.

\bibitem[{Lu et~al.(2024)Lu, Bansal, Xia, Liu, Li, Hajishirzi, Cheng, Chang, Galley, and Gao}]{lu2024mathvista}
Pan Lu, Hritik Bansal, Tony Xia, Jiacheng Liu, Chunyuan Li, Hannaneh Hajishirzi, Hao Cheng, Kai-Wei Chang, Michel Galley, and Jianfeng Gao. 2024.
\newblock Mathvista: Evaluating mathematical reasoning of foundation models in visual contexts.
\newblock In \emph{International Conference on Learning Representations (ICLR)}.

\bibitem[{Maaz et~al.(2023)Maaz, Rasheed, Khan, and Khan}]{maaz2023video}
Muhammad Maaz, Hanoona Rasheed, Salman Khan, and Fahad~Shahbaz Khan. 2023.
\newblock Video-chatgpt: Towards detailed video understanding via large vision and language models.
\newblock \emph{arXiv preprint arXiv:2306.05424}.

\bibitem[{Mayer et~al.(2025)Mayer, Ballout, Jassim, Nezami, and Bruni}]{mayer2025ivispar}
Julius Mayer, Mohamad Ballout, Serwan Jassim, Farbod~Nosrat Nezami, and Elia Bruni. 2025.
\newblock ivispar--an interactive visual-spatial reasoning benchmark for vlms.
\newblock \emph{arXiv preprint arXiv:2502.03214}.

\bibitem[{Piloto et~al.(2022)Piloto, Weinstein, Battaglia, and Botvinick}]{piloto2022intuitive}
Luis~S Piloto, Ari Weinstein, Peter Battaglia, and Matthew Botvinick. 2022.
\newblock Intuitive physics learning in a deep-learning model inspired by developmental psychology.
\newblock \emph{Nature human behaviour}, 6(9):1257--1267.

\bibitem[{Radford et~al.(2021)Radford, Kim, Hallacy, Ramesh, Goh, Agarwal, Sastry, Askell, Mishkin, Clark et~al.}]{radford2021learning}
Alec Radford, Jong~Wook Kim, Chris Hallacy, Aditya Ramesh, Gabriel Goh, Sandhini Agarwal, Girish Sastry, Amanda Askell, Pamela Mishkin, Jack Clark, et~al. 2021.
\newblock Learning transferable visual models from natural language supervision.
\newblock In \emph{International conference on machine learning}, pages 8748--8763. PMLR.

\bibitem[{Radford et~al.(2018)Radford, Narasimhan, Salimans, Sutskever et~al.}]{radford2018improving}
Alec Radford, Karthik Narasimhan, Tim Salimans, Ilya Sutskever, et~al. 2018.
\newblock Improving language understanding by generative pre-training.

\bibitem[{Raffel et~al.(2020)Raffel, Shazeer, Roberts, Lee, Narang, Matena, Zhou, Li, and Liu}]{raffel2020exploring}
Colin Raffel, Noam Shazeer, Adam Roberts, Katherine Lee, Sharan Narang, Michael Matena, Yanqi Zhou, Wei Li, and Peter~J Liu. 2020.
\newblock Exploring the limits of transfer learning with a unified text-to-text transformer.
\newblock \emph{Journal of machine learning research}, 21(140):1--67.

\bibitem[{Riochet et~al.(2021)Riochet, Castro, Bernard, Lerer, Fergus, Izard, and Dupoux}]{riochet2021intphys}
Ronan Riochet, Mario~Ynocente Castro, Mathieu Bernard, Adam Lerer, Rob Fergus, V{\'e}ronique Izard, and Emmanuel Dupoux. 2021.
\newblock Intphys 2019: A benchmark for visual intuitive physics understanding.
\newblock \emph{IEEE Transactions on Pattern Analysis and Machine Intelligence}, 44(9):5016--5025.

\bibitem[{Su et~al.(2023)Su, Lan, Li, Xu, Wang, and Cai}]{su2023pandagpt}
Yixuan Su, Tian Lan, Huayang Li, Jialu Xu, Yan Wang, and Deng Cai. 2023.
\newblock Pandagpt: One model to instruction-follow them all.
\newblock \emph{arXiv preprint arXiv:2305.16355}.

\bibitem[{Touvron et~al.(2023)Touvron, Lavril, Izacard, Martinet, Lachaux, Lacroix, Rozi{\`e}re, Goyal, Hambro, Azhar et~al.}]{touvron2023llama}
Hugo Touvron, Thibaut Lavril, Gautier Izacard, Xavier Martinet, Marie-Anne Lachaux, Timoth{\'e}e Lacroix, Baptiste Rozi{\`e}re, Naman Goyal, Eric Hambro, Faisal Azhar, et~al. 2023.
\newblock Llama: Open and efficient foundation language models.
\newblock \emph{arXiv preprint arXiv:2302.13971}.

\bibitem[{Vaswani et~al.(2017)Vaswani, Shazeer, Parmar, Uszkoreit, Jones, Gomez, Kaiser, and Polosukhin}]{vaswani2017attention}
Ashish Vaswani, Noam Shazeer, Niki Parmar, Jakob Uszkoreit, Llion Jones, Aidan~N Gomez, {\L}ukasz Kaiser, and Illia Polosukhin. 2017.
\newblock Attention is all you need.
\newblock \emph{Advances in neural information processing systems}, 30.

\bibitem[{Wang et~al.(2024)Wang, Chen, Han, Lin, Zhao, Liu, Zhai, Yuan, You, and Yang}]{wang2024exploring}
Yiqi Wang, Wentao Chen, Xiaotian Han, Xudong Lin, Haiteng Zhao, Yongfei Liu, Bohan Zhai, Jianbo Yuan, Quanzeng You, and Hongxia Yang. 2024.
\newblock Exploring the reasoning abilities of multimodal large language models (mllms): A comprehensive survey on emerging trends in multimodal reasoning.
\newblock \emph{arXiv preprint arXiv:2401.06805}.

\bibitem[{Weihs et~al.(2022)Weihs, Yuile, Baillargeon, Fisher, Marcus, Mottaghi, and Kembhavi}]{weihs2022benchmarking}
Luca Weihs, Amanda Yuile, Ren{\'e}e Baillargeon, Cynthia Fisher, Gary Marcus, Roozbeh Mottaghi, and Aniruddha Kembhavi. 2022.
\newblock Benchmarking progress to infant-level physical reasoning in ai.
\newblock \emph{Transactions on Machine Learning Research}.

\bibitem[{Yi et~al.(2020)Yi, Gan, Li, Kohli, Wu, Torralba, and Tenenbaum}]{yi2020clevrer}
Kexin Yi, Chuang Gan, Yunzhu Li, Pushmeet Kohli, Jiajun Wu, Antonio Torralba, and Joshua~B Tenenbaum. 2020.
\newblock Clevrer: Collision events for video representation and reasoning.
\newblock In \emph{International Conference on Learning Representations}.

\bibitem[{Yue et~al.(2024)Yue, Ni, Zhang, Zheng, Liu, Zhang, Stevens, Jiang, Ren, Sun, Wei, Yu, Yuan, Sun, Yin, Zheng, Yang, Liu, Huang, Sun, Su, and Chen}]{yue2023mmmu}
Xiang Yue, Yuansheng Ni, Kai Zhang, Tianyu Zheng, Ruoqi Liu, Ge~Zhang, Samuel Stevens, Dongfu Jiang, Weiming Ren, Yuxuan Sun, Cong Wei, Botao Yu, Ruibin Yuan, Renliang Sun, Ming Yin, Boyuan Zheng, Zhenzhu Yang, Yibo Liu, Wenhao Huang, Huan Sun, Yu~Su, and Wenhu Chen. 2024.
\newblock Mmmu: A massive multi-discipline multimodal understanding and reasoning benchmark for expert agi.
\newblock In \emph{Proceedings of CVPR}.

\bibitem[{Zhang et~al.(2022)Zhang, Van~Durme, Li, and Stengel-Eskin}]{zhang2022visual}
Chenyu Zhang, Benjamin Van~Durme, Zhuowan Li, and Elias Stengel-Eskin. 2022.
\newblock Visual commonsense in pretrained unimodal and multimodal models.
\newblock In \emph{Proceedings of the 2022 Conference of the North American Chapter of the Association for Computational Linguistics: Human Language Technologies}, pages 5321--5335.

\bibitem[{Zhang et~al.(2023)Zhang, Li, and Bing}]{zhang2023video}
Hang Zhang, Xin Li, and Lidong Bing. 2023.
\newblock Video-llama: An instruction-tuned audio-visual language model for video understanding.
\newblock \emph{arXiv preprint arXiv:2306.02858}.

\end{thebibliography}

\clearpage

\onecolumn
\appendix

\section{Intuitive Physics Results With Smaller Models}\label{subsec:vlm-results-small-models}
\begin{figure*}[th]
    \centering
    \begin{subfigure}[b]{\textwidth}
        \centering
        \includegraphics[width=\textwidth]{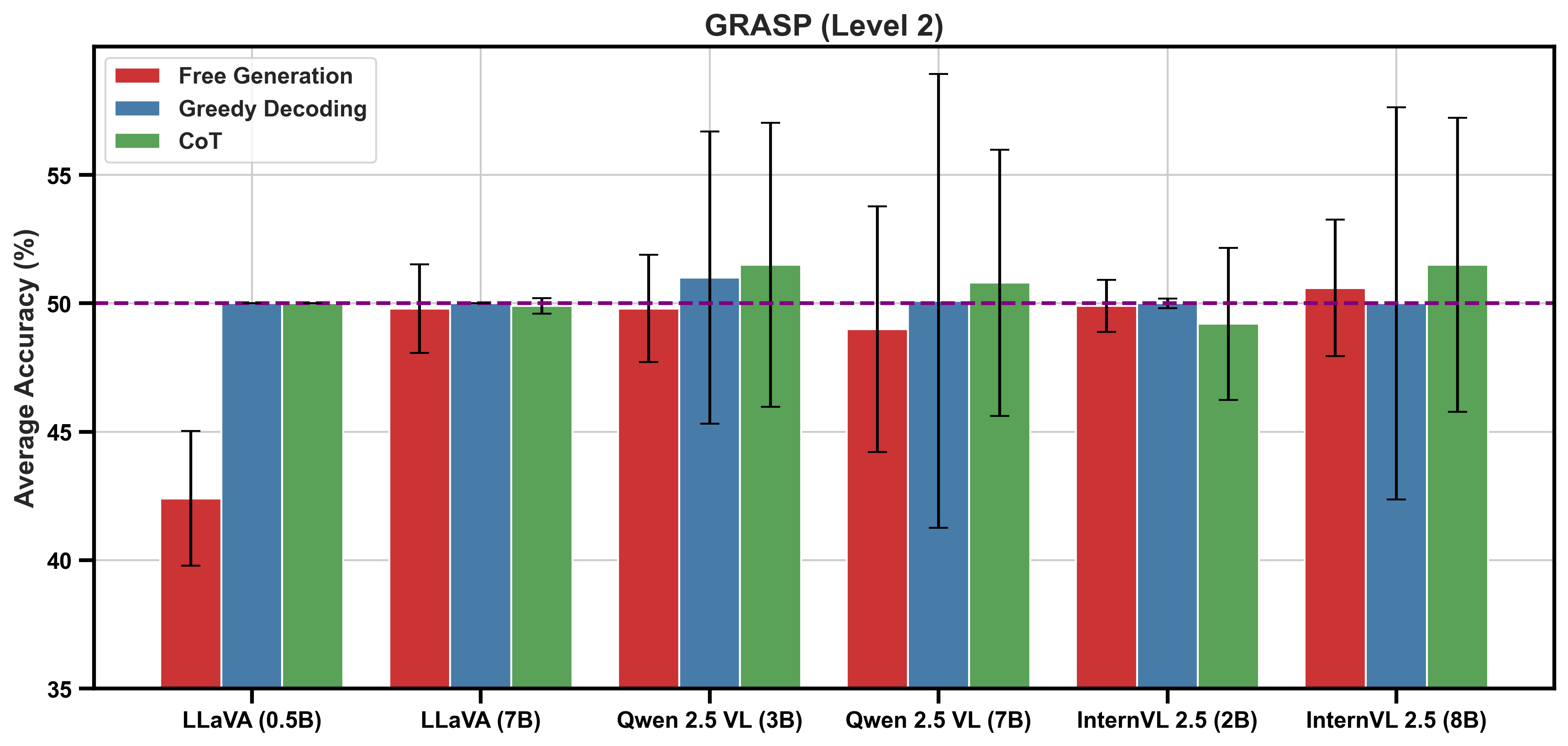}
    \end{subfigure}
    \vspace{3pt}

    \begin{subfigure}[b]{\textwidth}
        \centering
        \includegraphics[width=\textwidth]{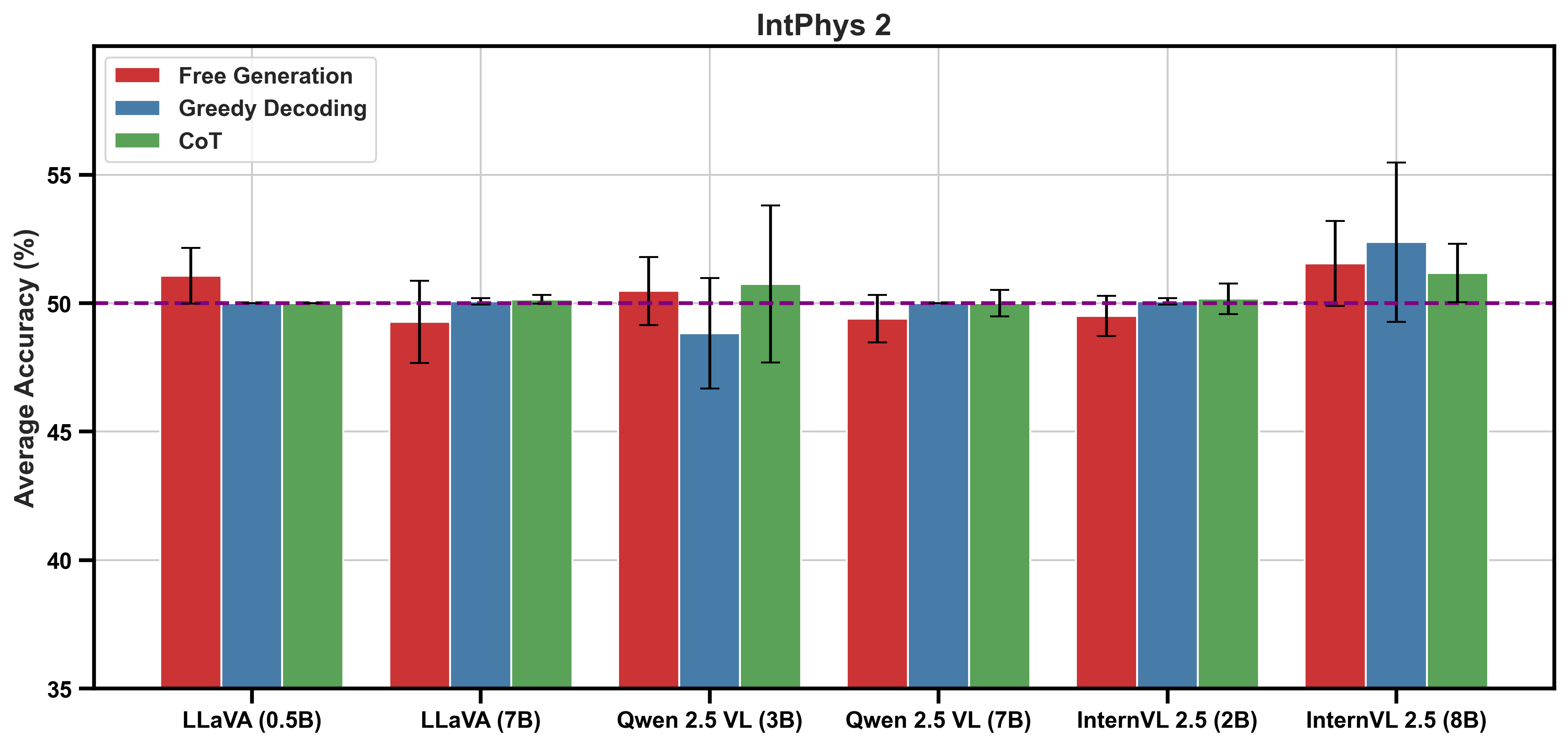}
    \end{subfigure}
    
    \caption{Average accuracy (\%) across the intuitive physics tests of GRASP and IntPhys 2 for the smaller sizes of the evaluated model families. The dashed line represents chance performance. Error bars denote the standard deviation across all tests of the respective dataset. All models perform close to chance.}
    \label{fig:vlm-results-small-models}
\end{figure*}

\clearpage

\section{Embedding Types and Dimensions}
\label{embed}
\begin{table*}[h!]
    \centering
    \scalebox{1.0}{
    \begin{tabular}{l|c|c|c}
        \toprule
        \textbf{Model Family} & \textbf{Parameters} & \textbf{Embedding Type} & \textbf{Embedding Dim.} \\
        \midrule
        \multirow{4}{*}{Qwen 2.5 VL} 
            & 3B, 7B, 72B & Vision (Before Proj.) & 1280 \\
            & 3B & Vision (After Proj.) or Lang. & 2048 \\
            & 7B & Vision (After Proj.) or Lang. & 3584 \\
            & 72B & Vision (After Proj.) or Lang. & 8192 \\
        \midrule
        \multirow{5}{*}{InternVL 2.5} 
            & 2B, 8B & Vision (Before \& After Proj.) & 1024 \\
            & 78B & Vision (Before \& After Proj.) & 3200 \\
            & 2B & Language Only & 2048 \\
            & 8B & Language Only & 4096 \\
            & 78B & Language Only & 8192 \\
        \midrule
        \multirow{4}{*}{LLaVA-OneVision} 
            & 0.5B, 7B, 72B & Vision (Before Proj.) & 1152 \\
            & 0.5B & Vision (After Proj.) or Lang. & 896 \\
            & 7B & Vision (After Proj.) or Lang. & 3584 \\
            & 72B & Vision (After Proj.) or Lang. & 8192 \\
        \bottomrule
    \end{tabular}
    }
    \caption{Comparison of embedding types and dimensions across model families. "Proj." refers to projection, "Lang." to language embeddings.}
    \label{tab:embedding-configs}
\end{table*}

\section{Qualitative Results}
\begin{table*}[h!]
    \centering
    \scalebox{0.8}{
    \begin{tabular}{lp{4in}}
         \toprule 
         \textbf{Video Descriptions} \\
         \midrule
         & \includegraphics[width=0.4\textwidth]{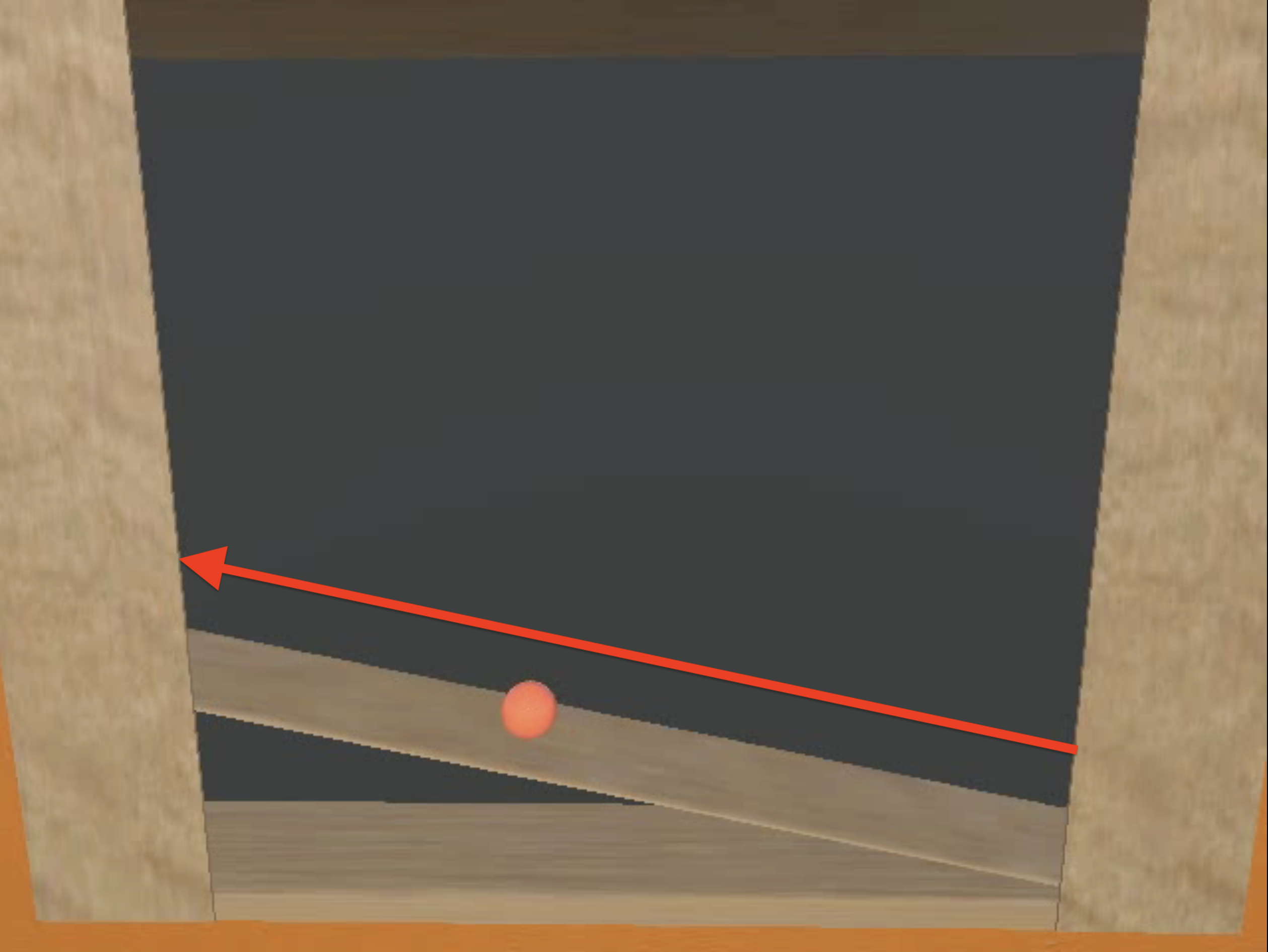} \\
         Prompt & What do you see in this video? \\
         \midrule
         LLaVA-OneVision (72B) & A wooden plank is seen inside a wooden frame against a black background. \\
         \midrule
         Qwen 2.5 VL (72B) & A small red ball rolls down a wooden ramp from the left side of the frame. It continues rolling until it exits the frame on the right side. The background is a dark gray, and the ramp is positioned at an angle, with a wooden wall on the left and right sides. The floor is a light orange color. \\
         \midrule
         InternVL 2.5 (78B) & I see a wooden door with a wooden frame. There is also a wooden plank leaning against the wall and a wooden door frame. \\
         \bottomrule
    \end{tabular}
    }
    \caption{Descriptions generated by the largest model variants for a video from the implausible Gravity scene in GRASP. The image shows one frame from the video, in which an orange ball rolls up a ramp—as indicated by the red arrow—without slowing down. All models exhibit some degree of hallucination in their responses.}
    \label{tab:descriptions}
\end{table*}

\clearpage

\section{Level 1 Prompts}
\label{prompt1}
\paragraph{Color} What color is the object on the table?
\paragraph{Directionality} Which direction is the ball rolling?
\paragraph{Movement} Is the ball moving?
\paragraph{Shape} What shape is the object on the table?

\section{Probing Prompts}
\label{prompt}
Following is the detailed prompt used for feature extraction from language models for level 2 of GRASP, which we refer to in the paper as the ``detailed prompt'':

Determine whether the video is physically plausible or implausible based on these concepts. Collision: plausible if a rolling ball hits another and both move per collision laws, implausible if they don’t move. Gravity: plausible if a dropped ball falls to the floor, implausible if it stops mid-air. Continuity: plausible if a ball rolls under an occluder and reappears in the correct position, implausible if it bypasses an obstacle illogically. Object Permanence: plausible if a ball under an
occluder reappears where expected, implausible if it’s in an unexpected spot or if an occluder flips fully despite an object behind it when it shouldn’t. Solidity: plausible if a ball stops at a wooden obstacle, implausible if it passes through. Unchangeableness: plausible if object colors remain consistent, implausible if they change. Gravity Continuity: plausible if a ball falls into a hole behind an occluder, implausible if it teleports past it. Gravity Inertia/Support: plausible if a falling or
pushed object lands on the ground, implausible if it hangs in mid-air. Inertia: plausible if a rolling ball bounces correctly off an edge or stops in the expected position, implausible if it bounces wrongly or appears on the opposite side. Determine plausibility by checking if the video aligns with physical laws.

Following is the simple prompt used for feature extraction from language models, which we refer to in the paper as the ``simple prompt'':
Is the video physically plausible or implausible?

\end{document}